\documentclass[11pt,a4paper]{article}

\usepackage{authblk}

\renewcommand\textsuperscript[1]{}

\usepackage{naaclhlt2018}
\usepackage{pgfplots}
\usepackage{rotating}
\usepackage{times}
\usepackage{latexsym}
\usepackage{tipa}
\usepackage{amsfonts}
\usepackage{amsmath}
\usepackage{amsthm}
\theoremstyle{plain}
\usepackage[small]{caption}

\usepackage{microtype}
\usepackage{graphicx}
\usepackage{subfigure}
\usepackage{url}
\usepackage{booktabs}
\usepackage{multirow}
\usepackage{adjustbox}

\usepackage{cleveref}
\crefname{section}{\S}{\S\S}
\Crefname{section}{\S}{\S\S}
\crefname{table}{Tab.}{}
\crefname{figure}{Fig.}{}
\crefname{algorithm}{alg.}{}
\crefname{equation}{eq.}{}
\crefname{appendix}{app.}{}
\crefformat{section}{\S#2#1#3}

\newcommand{\lemma}{\ell}
\newcommand{\lexeme}{\lamed}

\aclfinalcopy
\def\aclpaperid{45}

\newcommand{\word}[1]{\emph{#1}}

\newcommand{\vtheta}{{\boldsymbol \theta}}

\newcommand{\yy}{\mathbf{y}}
\newcommand{\xx}{\mathbf{x}}

\DeclareFontFamily{U}{rcjhbltx}{}
\DeclareFontShape{U}{rcjhbltx}{m}{n}{<->rcjhbltx}{}
\DeclareSymbolFont{hebrewletters}{U}{rcjhbltx}{m}{n}

\let\aleph\relax\let\beth\relax
\let\gimel\relax\let\daleth\relax

\DeclareMathSymbol{\aleph}{\mathord}{hebrewletters}{39}
\DeclareMathSymbol{\beth}{\mathord}{hebrewletters}{98}
\DeclareMathSymbol{\gimel}{\mathord}{hebrewletters}{103}
\DeclareMathSymbol{\daleth}{\mathord}{hebrewletters}{100}

\DeclareMathSymbol{\lamed}{\mathord}{hebrewletters}{108}
\DeclareMathSymbol{\mem}{\mathord}{hebrewletters}{109}
\DeclareMathSymbol{\ayin}{\mathord}{hebrewletters}{96}
\DeclareMathSymbol{\tsadi}{\mathord}{hebrewletters}{118}
\DeclareMathSymbol{\qof}{\mathord}{hebrewletters}{114}
\DeclareMathSymbol{\shin}{\mathord}{hebrewletters}{152}

\usepackage{xcolor}
\usepackage[disable]{todonotes}
\makeatletter
\newcommand*\iftodonotes{\if@todonotes@disabled\expandafter\@secondoftwo\else\expandafter\@firstoftwo\fi}
\makeatother
\newcommand{\noindentaftertodo}{\iftodonotes{\noindent}{}}

\newcommand{\jason}[2][]{{green!40}{#2}}

\newcommand{\Fixme}[2][]{\noindentaftertodo}

\newcommand{\Jason}[2][]{\noindentaftertodo}
\newcommand{\Ryan}[2][]{\noindentaftertodo}
\newcommand{\Chris}[2][]{\noindentaftertodo}
\newcommand{\Mans}[2][]{\noindentaftertodo}

\newlength{\extramargin}
\usepackage{calc}
\iftodonotes{\usepackage[paperwidth=\paperwidth+\extramargin*2,marginparwidth=\marginparwidth+\extramargin,width=\textwidth,height=\textheight]{geometry}}{}

\title{On the Diachronic Stability of Irregularity in Inflectional Morphology}

\author[*]{{\bf Ryan Cotterell}\raise1.0ex\hbox{\normalsize \textschwa}}
\author[*]{{\bf Christo Kirov}\raise1.0ex\hbox{\normalsize \textschwa}}
\author[]{{\bf Mans Hulden}\raise1.0ex\hbox{\normalsize \textipa{K}}}
\author[*]{{\bf Jason Eisner}\raise1.0ex\hbox{\normalsize \textschwa}}

\affil[*]{\raise1.0ex\hbox{\normalsize \textschwa}Department of Computer Science, Johns Hopkins University, Baltimore, MD,  USA}
\affil[]{\raise1.0ex\hbox{\normalsize \textipa{K}}Department of Linguistics, University of Colorado, Boulder, CO, USA}
\affil{{ \normalsize {\tt \{ryan.cotterell,ckirov1,eisner\}@jhu.edu}, {\tt mans.hulden@colorado.edu}}}

\begin{document}
\maketitle
\begin{abstract}
Many languages' inflectional morphological systems are replete with
irregulars, i.e., words that do not seem to follow standard inflectional
rules.
In
this work, we quantitatively investigate the conditions under which
irregulars can survive in a language over the course of time. Using recurrent neural networks to simulate language learners, we
 test the diachronic relation between frequency of words and their irregularity.
\end{abstract}

\section{Introduction}
When and why does irregularity persist? It would certainly be
easier to learn an exceptionless language. Nevertheless, irregularity
abounds at almost every level of language: some words fail to obey
otherwise universal phonological patterns, some verbs have irregular
conjugations, and some phrases have meanings that cannot be derived
compositionally. It is also clear, however, that systematicity is a
hallmark of human language---children learn regular rules that allow
them to analyze and produce novel utterances
\cite{pinker1994language}. Indeed, it is unlikely that a completely
unpredictable language could survive. In this paper, we explore the
limits of irregularity in the context of inflectional morphology. We
ask how many irregular inflections a language can possess and what
their distribution must be before a language becomes unlearnable and
regularized by the next generation. We employ neural
sequence-to-sequence models in a series of simulations
to this end.

The existence and degree of irregularity in inflectional morphology
remains a linguistic puzzle. English, for example, possesses a large
number of irregular verbs, ranging from full suppletion, e.g.,
\word{go} $\mapsto$ \word{went}, to no-longer-productive ablaut and umlaut patterns, e.g., \word{sing} $\mapsto$ \word{sang} and \word{fall} $\mapsto$ \word{fell}.
Recent work \cite{ackerman2013} has attempted
to explain irregularity, and, more generally,
morphological complexity, from a \emph{synchronic} perspective;
they make a typological claim about what sorts of languages
exist in terms of degree of irregularity.
The work presented here goes one step further beyond a snapshot
of extant languages---we
try to explain how language acquisition could allow irregulars over the course of
multiple generations. That is, we try to explain morphological
complexity from a \emph{diachronic} perspective. Concretely, we
conjecture the following: a morphological system will tend to retain
irregular forms only if they are ``sufficiently frequent.''
Inspired by the seminal work of \newcite{hare1995learning},
we investigate the claim with a series of neural simulations on
morphological data.

What makes our work different? Like \newcite{hare1995learning}, we focus on the diachronic
evolution of English verbal paradigms. However, rather than attempting to replicate the development from Old English to modern English, our experiments instead show that the distribution of irregularity in modern English is stable under generational transmission, compared to possible alternative distributions.
Furthermore, NLP's ability to generate inflected
morphological forms has greatly improved in recent years
\cite{cotterell-et-al-2017-shared}, both through the introduction of
recurrent neural models to the task and curation of new data. We take
advantage of these advances and offer a dramatically updated
simulation. In contrast to \newcite{hare1995learning}, our models
generate actual strings at each iteration, rather than
pseudo-phonological feature vectors termed Wickelphones
\cite{rumelhart1986learning}.

\paragraph{A Roadmap through the Paper.}
In the next two sections (\cref{sec:language-change} and
\cref{sec:inflection}), we briefly overview the language change and
acquisition literature and discuss our formalization of inflectional
morphology.
We then
progress into a discussion of the current state of the art in
neural-network modeling for inflection generation in
\cref{sec:neural-transducers}. Then, we introduce our simulation
scheme in \cref{sec:generational-modeling}, which allows us to show that most, though not all, infrequent irregulars are unsustainable as a language evolves.

\section{Language Change and Acquisition}\label{sec:language-change}
The principles that undergird diachronic language change
have fascinated linguists for more than a century. When comparing
intentional neologism, e.g.\ the formation of new technological
jargon, \newcite{paul1890principles} writes: ``the significance of
such capricious decisions is as nothing compared, with the slow,
involuntary and unconscious changes to which the [\ldots] usage of language
is perpetually exposed.'' The problem is still unsolved, but different
theories flourish. There
are two primary camps. The first, \textbf{acquisition-based change}, argues
that language is acquired imperfectly by
children and changes slowly over time. The second,
\textbf{usage-based change},
contrarily, argues that language changes continuously and involves gradual
adjustments over each speaker's life. We discuss each camp in turn
and relate them to the view here.

\paragraph{Acquisition-Based Change.}
In the generative tradition, linguistics has attributed much of
language evolution to its acquisition by children \cite{kroch2001syntactic}.
The manner in which children acquire irregular
morphology is one of the most-studied problems in
psycholinguistics. During acquisition, children in an English-speaking
environment display the following pattern. Initially, they apparently
memorize irregulars, correctly producing the past tense of \word{go}
as \word{went}. Afterwards, they overregularize, having picked up on the
fact that most English verbs add the ending /-d/.\footnote{More
  specifically, regular English verbs select among three allophones:
  [-d], [-t] and [-\textipa{1}d], depending on the previous phoneme.}
During this phase, children will produce \word{goed}, rather than
\word{went}. Finally, the child will recover, correctly producing the past
tense forms for both the regulars and irregulars in the lexicon. This
pattern has been widely observed and is termed \textbf{U-shaped
  learning} \cite{warren2012introducing}. What happens to rare forms, though? The argument then
continues that these rarer forms remain regularized, even if they were once
produced correctly as irregulars, such as the now-obsolete \word{shave} $\mapsto$ \word{shove}. Thus, only frequent forms tend to remain
irregular---this is well-attested
cross-linguistically \cite{lieberman2007quantifying}.

\paragraph{Usage-Based Change.}
Others, however, have argued that language change is continuous.  On this view, adult speakers are
constantly updating their internal linguistic representations as a
result of usage \cite{langacker1987foundations,bybee2006usage},
even though they have passed the so-called ``critical period'' of language development
\cite{lenneberg1967biological}.
One example of this comes from the domain of phonetics, where
\newcite{harrington2006acoustic} has argued that adult speakers modify
their speech patterns.  It is thus conceivable that some diachronic
language changes (perhaps even morphological ones) might be initiated
or abetted by adaptions in adult speakers, rather than having to wait
till the next generation.

\paragraph{The View in this Paper.}
The model in this paper assumes that all change is acquisition-based.
Following \newcite{hare1995learning}, we describe a series of
simulations, where one generation of a probabilistic model teaches the
next. In broad strokes, this is in line with the acquisition account
of change. The literature on the acquisition of morphology is rich, however,
and we caution that our simulations may not fully do justice to the process. For instance,
our simulations train only a single population model at each time step, whereas
language is learned and spoken by a community of individuals. Moreover, children
often start speaking before they have complete mastery of the tongue, as evinced
by the U-shaped pattern. Thus, we view our formulation and experiments as
a first pass at the problem, very much in line with other work in cognitive science, discussed in \cref{sec:recent-work-cogsci}.

\section{Inflectional Morphology Formalized}\label{sec:inflection}
We briefly outline our formalization for inflectional morphology
and thereby develop notation that we will use throughout the paper.
We adopt the framework of word-based morphology
\cite{aronoff1976word,spencer1991morphological}.
Thus, for the rest
of the work we define an \textbf{inflected lexicon} as a set of
word types.

Each word type is a triple of
\begin{itemize}
  \setlength\itemsep{0.5em}
\item a \textbf{lexeme} $\lamed$ (an arbitrary integer or string that indexes the
  word's core meaning and part of speech)
\item a \textbf{slot} $\sigma$ (an arbitrary integer or object that indicates how the
  word is inflected)
\item a \textbf{surface form} $f$  (a string over a fixed phonological or orthographic
  alphabet $\Sigma$)
\end{itemize}

We write $\pi(\lexeme)$ for the set of word types (triples) in the
lexicon that share the lexeme $\lexeme$, known as the {\bf paradigm} of
$\lexeme$.  The slots that appear in this set are said to be {\bf filled}
by corresponding surface forms.  For example, in the English paradigm
$\pi(\word{walk}_{\text{Verb}})$, the past-tense slot is filled by
\word{walked}. Often the lexeme $\lexeme$ is cited using its {\bf lemma},
which is the surface word $\ell$ associated with some particular slot
in $\pi(\lexeme)$; for example, in most languages, a verb's lemma is conventionally taken
to be its infinitive.
We note that nothing in our method requires a Bloomfieldian structuralist analysis
that decomposes each word into underlying morphemes: rather, this
paper is a-morphous in the sense of
\newcite{anderson1992morphous}.

More specifically, we will work within the UniMorph annotation scheme
\cite{sylak2016composition}.  In the simplest case, each slot
specifies a morpho-syntactic \textbf{bundle} of inflectional features
such as tense, mood, person, number, and gender.  For example, the
English surface form \word{walks} appears with a slot that indicates
that this word has the features $\left[\right.${\sc tense}$=${\sc
  present}, {\sc person}$=${\sc 3}, {\sc number}$=${\sc sg}$\left.\right]$.

However, in a language where two or more feature bundles
systematically yield the same form across all lexemes, UniMorph
generally collapses them into a single slot that realizes multiple
feature bundles.  Thus, a single ``verb lemma'' slot suffices to
describe all English surface forms in \{\word{see}, \word{go},
\word{jump}, \ldots\}: this slot indicates that the word can be a bare
infinitive verb, but also that it can be a present-tense verb that may
have any gender and any person/number pair other than
3rd-person/singular.

\section{Neural Transducers for Morphological Infection Generation}\label{sec:neural-transducers}
In the NLP literature, producing inflected forms given a lemma has become a
common task
\cite{durrett-denero:2013:NAACL-HLT,nicolai-cherry-kondrak:2015:NAACL-HLT,ahlberg-forsberg-hulden:2015:NAACL-HLT,faruqui-EtAl:2016:N16-1,cotterell-EtAl:2016:SIGMORPHON}. It
generally involves learning a string-to-string mapping with (often)
monotonic alignments between the characters. This is the NLP
community's analogue to the past-tense generation task originally
considered by \newcite{rumelhart1986learning}. The goal is to train a
model capable of mapping the lemma (in the case of English, the stem)
to each form in the paradigm. In the case of English, the goal would
be to map a lemma, e.g., \word{walk}, to its past tense word
\word{walked} and to its gerund \word{walking} and 3$^\text{rd}$  person
present singular \word{walks}.

The state of the art on this task is currently held by an
encoder-decoder recurrent network
\cite{cotterell-EtAl:2016:SIGMORPHON}. This architecture consists of two
LSTM \cite{hochreiter1997long} recurrent neural networks (RNNs)
coupled together by an attention mechanism. The encoder RNN reads each
symbol in the input string one at a time, first assigning it a unique
embedding, then processing that embedding to produce a representation
of the phoneme given the rest of the phonemes in the string. The
decoder RNN produces a sequence of output phonemes one at a time,
using the attention mechanism to peek back at the encoder states as
needed. Decoding ends when a halt symbol is output. Given $\xx, \yy \in \Sigma^*$, the
encoder-decoder architecture encodes the probability distribution over forms
\begin{align}
  p(\yy \mid \xx) &= \prod_{i=1}^N p(y_i \mid y_1, \ldots, y_{i-1}, c_i) \\
                     &= \prod_{i=1}^N g(y_{i-1}, s_i, c_i),
\end{align}
where $g$ is a non-linear function (in our case it is a multi-layer perceptron), and $s_i$
is the hidden state of the decoder RNN.
Note that we write $\xx = (x_1, \ldots, x_M)$, where $x_i \in \Sigma$, and $\yy = (y_1, \ldots, y_{N})$, where $y_i \in \Sigma$.
Finally, $c_i$ is a convex combination of the the encoder RNN hidden states $h_{i}$,
using the attention weights $\alpha_{k}(s_{i-1})$ that are computed based on the previous decoder hidden state: $c_i = \sum_{k=1}^{|\xx|} \alpha_{k}(s_{i-1}) h_{k}$.
We refer the reader to \newcite{DBLP:journals/corr/BahdanauCB14} for the complete architectural specification of the model.

The above formulation works best in the case of a string-to-string translation. However,
the inflection task is more accurately described as a \textbf{labeled transduction} problem.
Specifically, we would like to produce a different output depending on an additional
label. In our case, this is the slot $\sigma$. To give a concrete example, we would like
to transduce the English lemma \word{walk} to \word{walked} if we condition
on the slot $\sigma = \left[\text{\sc tense} = \text{\text{\sc past}} \right]$,
but map \word{walk} to \word{walking} if we condition on the slot $\left[\text{\sc tense} = \text{\text{\sc gerund}} \right]$.
In the labeled inflection scenario, we define $\yy = f$ and $\xx$ is a concatenation of $\sigma$
and $\ell$. If we consider the morphological features to be taken from an alphabet $\Delta$,
we can then feed in a string $\text{\sc concat}\left(\mathbf{\sigma}, \ell \right) \in \Delta^* \Sigma^*$. In English,
$\Delta = \{ \text{\sc tense} = \text{\sc past}, \text{\sc tense} = \text{\sc gerund}, \ldots \}$.
As an actual example, consider the source string {\footnotesize {\tt GERUND w a l k} }
and target string {\footnotesize {\tt w a l k i n g}}.
    This encoding procedure is described in detail
in \newcite{kann-schutze:2016}.

\section{Generational Modeling}\label{sec:generational-modeling}
We focus on a paradigm called \textbf{generational learning}
\cite{hare1995learning}, wherein we will examine the ability of neural
models to convey their learned linguistic knowledge to other models
with the intent of simulating how language is passed on from
generation to generation.  Our simulations focus on the transmission
of knowledge of inflectional morphology in idealized conditions. In
our simulations, we define a series of \textbf{production} models that
will produce inflections, given a lemma and a slot as input. Each of
these will be formulated as an LSTM sequence-to-sequence model, as
discussed in \cref{sec:neural-transducers}. At each generation, we
train the production model off of output samples from the previous
generation (with the exception of the first generation, where
we train using gold output forms instead). Thus, the previously trained
model \emph{teaches} the next generation, simulating, albeit somewhat
crudely, how language is passed along over time.

\begin{figure*}
  \includegraphics[width=1.0\textwidth]{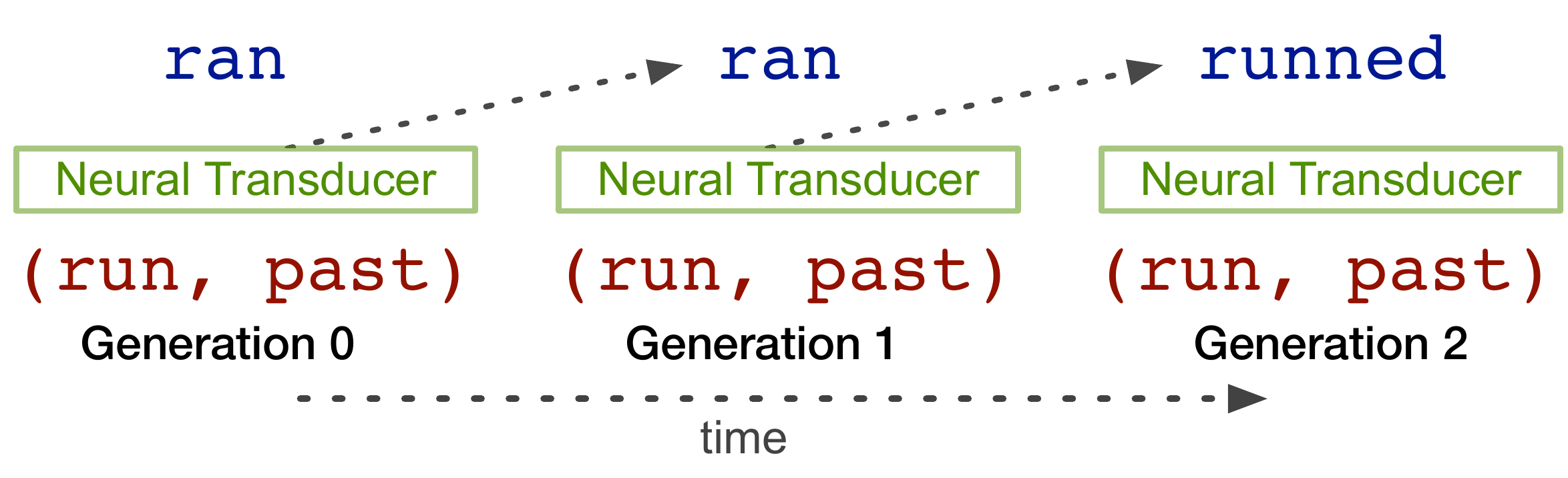}
  \caption{Graphical depiction of our generation learning scheme. At each generation, a form is sampled from the previous generation to retrain
  the neural transducers. In this toy example, we have that the irregular \word{run}$\mapsto$\word{ran} is conserved in the first time step, but
  regularizes to \word{run} $\mapsto$ \word{runned} after two steps.}
  \label{fig:generational-simulation}
\end{figure*}

\paragraph{The Production Model $p_{\vtheta}(f \mid \lemma, \sigma)$.}
Our production model takes the form of a conditional distribution $p_{\vtheta}(f
\mid \lemma, \sigma)$, which is parametrized as an LSTM-based
sequence-to-sequence model with attention as described in
\cref{sec:neural-transducers}.  We interpret $p_{\vtheta}(f
\mid \lemma, \sigma)$ as a distribution over possible productions the
\emph{population} may emit when attempting to inflect the lemma
$\lemma$ for the slot $\sigma$. We emphasize the population, because an
individual is relatively deterministic in how they produce a
form---few adult speakers alternate between \word{goed} and
\word{went} as the past tense of \word{go} in normal speech, with the
exception of occasional speech errors. Instead, at the population
level, we interpret the probability as the
percentage of the population that utters each string in $\Sigma^*$ as
the target inflection.\footnote{We emphasize again that this only one way
of effecting a generational learning scheme. For instance, a multi-agent model may be more appropriate in that different learners may incorporate input from different sources.}

\paragraph{Generational Simulation.}
Now, given our production model, we describe a procedure for
performing a generational simulation, wherein at each generation, a
new network learns from the previous generation.  This process is shown in \cref{fig:generational-simulation}.  We simulate $T$
generations, repeatedly sampling from a distribution $q$ over inputs
$(\lemma,\sigma)$. We will discuss various choices of $q$ in
\cref{sec:experimental-design}.

At generation $t=0$, our inflected lexicon consists of the {\em true} set of English triples $(\lemma,\sigma,f^{(0)})$.

At each generation $1 \leq t \leq T$, we train morphological parameters $\vtheta^{(t)}$ to maximize $\mathbb{E}_q[\log p_{\vtheta^{(t)}}(f^{(t-1)} \mid \lemma, \sigma)]$.  For this, we run stochastic gradient descent for 100,000 iterations.  Thus, at each iteration we draw some $(\lemma,\sigma) \sim q$ and adjust the parameters to increase the conditional probability of the corresponding form $f^{(t-1)}$ given $(\lemma,\sigma)$.

Then, for each triple $(\lemma,\sigma,f^{(t-1)})$ in the lexicon---whether or not it was ever used for training---we replace the form $f^{(t-1)}$ with a random sample $f^{(t)}$ drawn from the newly trained distribution $p_{\vtheta^{(t)}}(\cdot \mid \lemma, \sigma)$.  This gives us our new inflected lexicon at generation $t$.

We can view this process as simulating a Markov chain with a very
elaborate transition procedure---one in which we have to train a
neural network for a given number of epochs to find the next
state. This is in line with previous work that simulates language
change through Markov chain modeling \cite{niyogi2009proper}.

\paragraph{The Role of Regularization.}
Our production models are explicitly regularized using both early stopping and dropout. The finite size of the neural networks used is also a form of regularization. In machine learning parlance, this is
done to generalize to held-out data. While we typically evaluate our
models on held-out data, in our multi-generational setting we are mainly
interested in how each subsequent model changes its predictions on the \emph{
  training} data over time. Why? While most NLP papers
  are focused on generalization to new data, we are concerned with
  how the most frequent words in the lexicon evolve. Thus, we consider
  how a regularized learner performs on the \emph{training} data.
   As a real-world example of such change, consider the past tense of the
relatively frequent regular English verb \word{bake}:
\word{baked}. The verb is of Germanic stock, derived from the Old
English verb \word{bacan} with irregular past tense \word{boc} \cite{lass1994old}. In
modern orthography, \word{boc} would correspond to \word{boke}. When
modeling the evolutionary change of language, we need an external
pressure to force the model to create general rules for its morphology rather than simply memorizing every mapping and passing it on untouched to the next generation. Neither synchronic generalization nor diachronic change would be possible without the kind of pressure that regularization applies.

\section{Irregularity and Frequency}\label{sec:irregularity-frequency}
In the previous sections we have discussed a formalization of
inflectional morphology (discussed in \cref{sec:inflection}) and the development of a simulation scheme
for the evolution of inflectional morphology over time (discussed in
\cref{sec:generational-modeling}). How does it all piece together?
Our interest in simulation rests on our desire to
attempt to provide evidence for the following conjecture about
language change. Each of the previous sections describes a step towards that goal.

\newtheorem*{theorem*}{Conjecture}
\begin{theorem*}
  Over generations, irregular inflectional forms are only able to remain
   in a language if they appear with sufficient
  frequency, or are phonologically or morphologically similar to frequent irregular forms (in other words, if they follow a very common irregular pattern).
  For example, the past tense of \word{undergo} is unlikely to regularize
  to \word{undergoed} because speakers will associate it with the extremely common pattern of \word{go} $\mapsto$ \word{went}.\footnote{ One interesting possible exception to this claim is the
    case of the \textit{pass{\'e} simple} in French, where
    forms---often irregular---that are very uncommon in the spoken
    vernacular, nevertheless persist in the written form, exhibiting a
    clear case of diglossia. We believe, however, that without the
    intervention of the Francophone school system, these forms would
    naturally die out as they have in other Romance languages such as
    Romansh \cite{beninca2005rhaeto}.}
\end{theorem*}

The conjecture above has been repeated in a number of places in the
scientific literature.  For example, \newcite{hare1995learning} write
``[o]ur claim is that this regularization process resulted from the
difficulty in learning of items that had neither high type frequency
nor phonological class cohesion to support them.'' In other words,
they argue that infrequent, irregular verbs should be hard to
learn. Likewise, \newcite{dowman2006innateness} write ``put simply, frequent
verbs can afford to be irregular, since they will have ample opportunity to be
transmitted faithfully through the bottleneck \cite{kirby2001spontaneous}.''

What distinguishes our approach? In our view, previous work had the shortcoming of not being able to provide an explicit
distribution over sequences in $\Sigma^*$.  In the last two years, NLP
has made large advances in morphological inflection generation and,
more generally, sequence-to-sequence transduction tasks such as machine
translation. The work of \newcite{hare1995learning}, which used
the model of \newcite{rumelhart1986learning}, relies on predicting
binary feature vectors associated with strings, rather than
the strings themselves. By contrast, we directly
parameterize a distribution over all of $\Sigma^*$.
Thus, our model is capable of sampling full strings
at each generation. In short, we believe our work
gets the NLP aspect of this cognitive problem right.

\section{Experimental Design}\label{sec:experimental-design}
The goal of our simulation experiments is to show that infrequent
irregular forms in a language will eventually regularize. The
experimental variable in this simulation is, thus, the unigram
distribution over types.

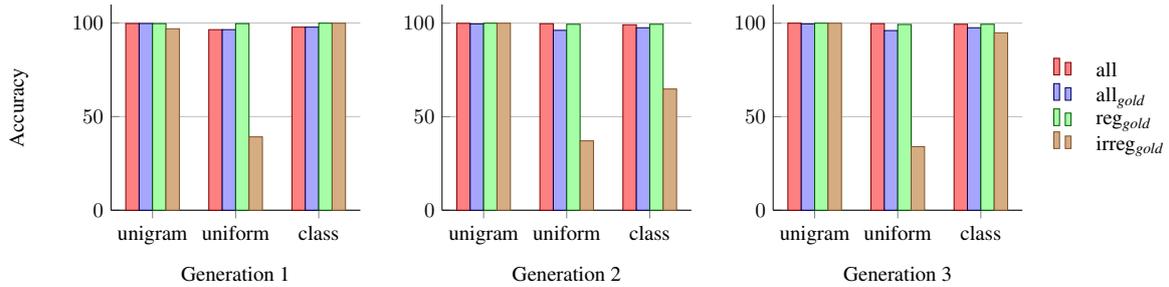
\begin{figure*}
 \centering
\begin{adjustbox}{width=2\columnwidth}
  \begin{tikzpicture}
    \begin{axis}[
        axis x line*=bottom, axis y line*=left,
        ybar=0pt,
        ymin=0,
        ylabel={Accuracy},
        y label style={at={(axis description cs:-0.1,0.5)}},
        xlabel={Generation 1},
        x label style={at={(axis description cs:0.5,-0.1)}},
        bar width=7pt,
        ymajorgrids,
        yminorgrids,
        symbolic x coords={unigram,uniform,class},
        enlarge x limits={value=0.25, auto},
        xtick=data,
        nodes near coords align={horizontal}, every node near coord/.append style={rotate=90},
        width=16em, height=14em,
        legend style={at={(1.1,0.5)},anchor=west,draw=none,fill=none,column sep=1em,name=legend},
        legend columns=1,
        legend cell align=left
        ]
      \addplot[  red!50!black,fill=  red!50!white] coordinates {(unigram, 99.72) (uniform, 96.43) (class, 97.77)};
      \addplot[ blue!35!black,fill= blue!35!white] coordinates {(unigram, 99.72) (uniform, 96.43) (class, 97.77)};
      \addplot[green!35!black,fill=green!35!white] coordinates {(unigram, 99.65) (uniform, 99.65) (class, 99.88)};
      \addplot[brown!65!black,fill=brown!65!white] coordinates {(unigram, 96.91) (uniform, 39.18) (class, 99.88)};
    \end{axis}
  \end{tikzpicture}
  \hspace{1.2em}
  \begin{tikzpicture}
    \begin{axis}[
        axis x line*=bottom, axis y line*=left,
        ybar=0pt,
        ymin=0,
        xlabel={Generation 2},
        x label style={at={(axis description cs:0.5,-0.1)}},
        bar width=7pt,
        ymajorgrids,
        yminorgrids,
        symbolic x coords={unigram,uniform,class},
        enlarge x limits={value=0.25, auto},
        xtick=data,
        nodes near coords align={horizontal}, every node near coord/.append style={rotate=90},
        width=16em, height=14em,
        legend style={at={(1.1,0.5)},anchor=west,draw=none,fill=none,column sep=1em,name=legend},
        legend columns=1,
        legend cell align=left
        ]
      \addplot[  red!50!black,fill=  red!50!white] coordinates {(unigram, 99.92) (uniform, 99.72) (class, 99.15)};
      \addplot[ blue!35!black,fill= blue!35!white] coordinates {(unigram, 99.63) (uniform, 96.19) (class, 97.56)};
      \addplot[green!35!black,fill=green!35!white] coordinates {(unigram, 100.0) (uniform, 99.42) (class, 99.42)};
      \addplot[brown!65!black,fill=brown!65!white] coordinates {(unigram, 100.0) (uniform, 37.11) (class, 64.82)};
    \end{axis}
  \end{tikzpicture}
  \hspace{1.2em}
  \begin{tikzpicture}
    \begin{axis}[
        axis x line*=bottom, axis y line*=left,
        ybar=0pt,
        ymin=0,
        xlabel={Generation 3},
        x label style={at={(axis description cs:0.5,-0.1)}},
        bar width=7pt,
        ymajorgrids,
        yminorgrids,
        symbolic x coords={unigram,uniform,class},
        enlarge x limits={value=0.25, auto},
        xtick=data,
        nodes near coords align={horizontal}, every node near coord/.append style={rotate=90},
        width=16em, height=14em,
        legend style={at={(1.1,0.5)},anchor=west,draw=none,fill=none,column sep=1em,name=legend},
        legend columns=1,
        legend cell align=left
        ]
      \addplot[  red!50!black,fill=  red!50!white] coordinates {(unigram, 100.0) (uniform, 99.76) (class, 99.43)};
      \addplot[ blue!35!black,fill= blue!35!white] coordinates {(unigram, 99.63) (uniform, 96.06) (class, 97.56)};
      \addplot[green!35!black,fill=green!35!white] coordinates {(unigram, 100.0) (uniform, 99.31) (class, 99.42)};
      \addplot[brown!65!black,fill=brown!65!white] coordinates {(unigram, 100.0) (uniform, 34.02) (class, 94.86)};
      \legend{all,all$_{\textit{gold}}$,reg$_{\textit{gold}}$,irreg$_{\textit{gold}}$}
    \end{axis}
  \end{tikzpicture}
\end{adjustbox}
\caption{Here, we present accuracy of our model in the first three generations. The red \textit{all} bars indicate how well each generation learns to predict the \emph{previous} generation's output (even if that output is no longer gold).  By contrast, the blue all$_{\textit{gold}}$ bars indicate accuracy as measured against the original morphological lexicon.  Thus, if a form regularizes in generation 2, \textit{all} credit would be given in generation 3 only for predicting the regular form, while all$_{\textit{gold}}$ credit would be given only for returning to the original irregular.  The bars reg$_\textit{gold}$ and irreg$_\textit{gold}$ break down the all$_\textit{gold}$ bars into a hand-annotated regular and irregular dichotomy, provided by \newcite{albright2003rules}.  We see that verb production stabilizes over time.  After an initial drop in the ability to correctly predict the irregulars at generation 1, the remaining irregulars continue to be passed on unregularized to the following generations.  Note that the regular-irregular classification is taken from \newcite{albright2003rules} and does not cover the full training set. Thus, it is not a complete break-down.\jason{some results look fishy, e.g., reg gold and irreg gold are both better then all gold in generation 1 class}}
\label{fig:accuracy}
\end{figure*}

\subsection{Stability under Unigram Distributions}

How stable is the irregular system in English?
We investigate several different distributions $q$ over inputs
(lemma-slot pairs). We contend that if the distribution $q$ does
not put sufficient probability on the inputs that map to irregular forms, those
forms will regularize over time. We consider three frequency distributions.

\paragraph{True Unigram Distribution.}
First, we consider the unigram distribution extracted from the Google $n$-gram database
\cite{brants2006web}.  As it is typical for irregulars to be much more
frequent than regulars in modern languages, this samples from this
distribution will often be irregular.
We will refer to this distribution as \textbf{unigram}
henceforth.

\paragraph{Uniform Distribution over Types.}
Next, we consider the uniform distribution over types. This distribution treats all
forms equally. Thus, irregulars will not be very common. We expect the
model to struggle to learn irregulars in this situation, picking up the
regular patterns much faster. We will refer to this distribution as \textbf{uniform} henceforth.

\paragraph{Teasing Apart Word Classes.}
Finally, we consider a special type of probability permutation
$\rho$ that will tease apart the fact that some
frequent phonologically and morphologically irregular forms may prop
up less frequent irregulars (``gang effects''). It works as follows.

Suppose our training data consists of a set of complete paradigms for
the same part of speech (in our case, verbs), so that the lexemes
$\lexeme$ are taken from a set $L$, the slots $\sigma$ are taken from
a set $S$, and all forms that realize pairs in $L \times S$ are
included in the dataset.

We now select
$L_\textit{test}$, a random 20\% of $L$, and $S_\textit{test}$, a random 40\% of $S$.  We choose a random
permutation $\rho_\lexeme$ of $L$ that preserves the set $L_\textit{test}$.
and a random permutation $\rho_{\sigma}$ of $S$ that preserves the set $S_\textit{test}$.
Now, we can define the function
\begin{align}
\rho(\lexeme, \sigma) &= (\lexeme, \sigma)\,\,\,\,\,\,\text{if }\lexeme \in L_{\textit{test}}\text{ and }\sigma \in S_{\textit{test}} \nonumber\\
\rho(\lexeme, \sigma) &= \left(\rho_\lexeme(\lexeme), \rho_\sigma(\sigma)\right)\,\,\,\,\,\,\text{otherwise}
\end{align}
$\rho$ is now itself a permutation that preserves
all pairs in $L_\textit{test} \times S_\textit{test}$, but scrambles
the others.

Each word type $w \in W$ has some $(\lexeme, \sigma)$ pair, and we
replace its frequency with the frequency of the word whose pair is
instead $\rho(\lexeme, \sigma)$.  If our dataset contains collections
of paradigms for several parts of speech, we can run this frequency
permutation on each part of speech separately.

We can now run our experiment with the unigram distribution from this
modified lexicon, which changes frequencies but not forms.  In
particular, each lexeme $\lexeme$ still has the same lemma and the
same paradigm.  Only the sampling distribution $q$ has changed.

Note that $\rho$ preserves the frequencies of some
word types.  Under this experimental design, we only evaluate on those
word types.  This is because we want to check how inputs with their
\emph{original, unaltered} frequencies are affected by permuting the
frequencies of the {\em rest} of the verbs in the system, affecting
phonological cohesion in the process.

\begin{table*}
 \centering
 \begin{adjustbox}{width=2\columnwidth}
  \begin{tabular}{ll llll llll llll}  \toprule
    && \multicolumn{4}{c}{Generation 1} & \multicolumn{4}{c}{Generation 2} & \multicolumn{4}{c}{Generation 3}  \\ \cmidrule{3-14}
      && all & all$_{\textit{gold}}$ & reg$_{\textit{gold}}$ & irreg$_{\textit{gold}}$ & all & all$_{\textit{gold}}$  & reg$_{\textit{gold}}$ & irreg$_{\textit{gold}}$ & all & all$_{\textit{gold}}$  & reg$_{\textit{gold}}$ & irreg$_{\textit{gold}}$ \\ \cmidrule{1-2} \cmidrule(lr){3-6} \cmidrule(lr){7-10} \cmidrule(lr){11-14}
     & unigram &  99.72\% & 99.65\% & 99.72\%  & 96.91\% & 99.92\% & 99.63\% & 100.0\% & 100.0\% & 100.0\% & 99.63\% & 100.0\% & 100.0\% \\
       \begin{rotate}{90}0/1\end{rotate}   & uniform &  96.43\% & 96.43\% & 99.65\% & 39.18\% & 99.72\% & 96.19\% & 99.42\% & 37.11\% & 99.76\% & 96.06\% & 99.31\% & 34.02\%\\
   & class &  97.77\% & 97.77\% & 99.88\% & 95.32\% & 99.15\% & 97.56\% & 99.42\% & 94.32\% & 99.43\% & 97.56\% & 99.42\% & 93.95\% \\ \cmidrule{2-14}
     & unigram & -1.13  & -1.49 & -1.13 & -11.34 & -0.32 & -8.72 & -8.16 & -107.29 & -0.34 & -11.92 & -11.34 & -129.72  \\
       \begin{rotate}{90}$D_{\textit{KL}}$\end{rotate}   & uniform & -30.63 & -30.63 & -0.41 & -424.21 & -0.23 & -239.17 & -8.79 & -4288.35 & -0.24 & -254.27 & -14.74 & -4503.88 \\
   & class &  -40.83 & -40.83 & -0.54 & -616.61 &-0.53 & -78.54 & -4.84 & -1357.41 & -0.32 & --85.43 & -6.78 & -1478.45 \\
    \bottomrule
  \end{tabular}
  \end{adjustbox}
  \caption{Here we report the performance of the neural transducer under the two metrics across generations. All models were trained for $100$ epochs. We see that under all three distributions, predictions stabilize in the sense that later generations learn to replicate the outputs of previous generations without applying any additional changes to verbs such as regularization of irregulars. Note that the regular-irregular classification is taken from \newcite{albright2003rules} and does not cover the full training set. Thus, it is not a complete break-down.}
  \label{tab:results}
\end{table*}

\subsection{Evaluation}
We consider two evaluation metrics when comparing learning
under the different distributions $q$. These metrics
take the form of test statistics over the training data. Using
a paired permutation test, we attempt to reject the null hypothesis
that the following test statistics are the same under different distributions $q$: (i) the number of forms that have changed after $t$ generations and (ii) the log-probability of the original forms under the degraded model, i.e., $\text{KL}\left(p^{(0)} \mid \mid p^{(t)}\right)$. The first
metric tells us how accurate the model remains at predicting the original inflections in the training corpus. This
is the standard evaluation metric for morphological generation. The second metric is softer---it can give partial credit
if the original form is not \emph{that} unexpected even after $t$ generations.

\subsection{Early Stopping as Regularization}
The neural transducer we employ in this study is sufficiently
high-capacity so as to ensure it can memorize the training
data. Thus, in order for our simulation
to be a success, we must prevent the network from simply memorizing
all the data---this would not give the forms an opportunity
to evolve. However, empirical work
has suggested that neural networks prioritize learning simpler
patterns in the data first \cite{arpit2017closer}. Thus,
we opt to stop the learner before it has converged \cite{goodfellow2016deep}.
Each generation's model is trained for $E=100$ epochs.  We also
employed drop-out on the recurrent layers with a dropout probability
of $0.3$. All experiments are performed
with the open-source toolkit OpenNMT \cite{klein2017opennmt}.

\subsection{Experimental Data}
Following \newcite{hare1995learning}, we focus on verbal inflection in
English.  The data are taken from the {\sc UniMorph} collection. English is
of Indo-European stock, from the Germanic branch.  Its verbal
inflection is modest, distinguishing 5 forms, typically. However, it
has a large collection of irregular verbs.  To create the experimental
data, we first took 4039 past tense forms, selected by
\newcite{albright2003rules}.  For each of those forms that has an
entry in {\sc UniMorph}, we expand it into its full paradigm.

\subsection{Model Parameters}

As described in \cref{sec:neural-transducers}, we used encoder-decoder architectures with global attention. Specifically, both the encoder and decoder consisted of 2-layer LSTMs. The encoder was bidirectional and output from the forward and backward LSTMs was concatenated. Both encoder and decoder had 100 hidden units and all character embeddings were 300 hidden units.  Networks were trained using Adadelta with a base learning rate of 1.0. Minibatches of size 20 were used. Dropout between layers was set at 0.3.

\section{Discussion and Analysis}
We find with $p < 0.002$ that \emph{all} pairwise differences between the
uniform, unigram and class distributions are
significant under a paired permutation test.\footnote{To compare two
  models, we swap their predictions using a randomly generated
  permutation. The computation of the $p$-value is described at \url{http://axon.cs.byu.edu/Dan/478/assignments/permutation_test.php}.}$^,$\footnote{In the case of the class distribution, we only compare on the $F_\textit{test}$ set to keep the test paired.} Moreover, the size of the difference is quite large:
while 96.91\% of the irregulars are memorized under the true unigram
distribution after 100 epochs, only 39.18\% are memorized under our artificial uniform distribution. We see
clearly in \cref{fig:accuracy} that the transition stabilizes. In
short, the regularized changes that happen during the transmission
from generation 1 to 2 are then maintained and ossified during the
transmission from generation 2 to 3. Thus, under the neural transducer
model, if a language finds itself in an unstable condition with
respect to the frequency distribution over its irregulars, it will
attempt to regularize it---infrequent forms will become more
regular. To show that these forms are actually regularized, we
present a smattering of randomly sampled mistakes the neural
transducer made after the first generation, shown in
\cref{tab:model-output}. As expected, we find that
the uniform distribution performs significantly worse than the other two
in terms of performance with respect to the gold.

\paragraph{The Role of Phonological Cohesion.}
Our original conjecture stated that less frequent irregulars can be
retained in a language if they are part of a pattern---e.g.,
\word{underwent} gets support from \word{went} and also from
\word{undergone}.  Under the class distribution, we show evidence for
this.  We evaluate on irregulars with unpermuted frequency, i.e.,
$F_\textit{test}$, while destroying the frequencies of related words
through a random permutation. If frequency were all that mattered in
preserving the continuation of irregularity, we would expect these
irregulars to decay \emph{at the same rate} as the those under the
unigram distribution. As evinced in \cref{tab:results}, this is not
the case. Irregulars decay at a somewhat faster rate when the
frequencies of \emph{other} forms are randomly permuted.  This
suggests that irregular forms in modern English may have survived in part
thanks to the {\em actual} frequencies of other forms.

\paragraph{What does the simulation tell us?}
We find that the actual ability to transfer irregularity between
generations depends on the frequency distribution. Neural transducers
seem to pick up the general pattern of regular verbs long before they
master individual irregulars.  Of course, this is to be expected as
\emph{most} verbs are, in fact, regular. So, in answer to the original
question we posed, neural learners can still manage to assimilate
irregular patterns if they are frequent enough. Naturally, the
modeling assumptions of the neural networks do not perfectly accord with
what we know about human learners.

The class permutation experiment also gives us insight into how
related irregulars interact.  It seems that the network's ability to
learn relatively infrequent irregular forms, such as \word{underwent},
hinges in part on the frequency of related irregular forms such as
\word{went}.  To our knowledge, this work is the first computational
experiment that focuses on the actual generation of complete word
forms, in contrast to \newcite{hare1995learning}, that shows such
behavior in a simulation.\footnote{Our significance test only shows
  that \emph{this specific} permutation underperforms compared to the
  true unigram distribution. A better experimental design would be to
  run thousands of permutations to allow us to test whether the
  unigram distribution is actually better than an arbitrary
  permutation of this class.  We would also like to run experiments
  that take $S_{\textit{test}}=S$ or $L_{\textit{test}}=L$, to tease
  apart the effect of high-frequency related lemmata for the same slot
  (e.g., (\word{undergo}, {\sc past}) $\mapsto$ \word{went} is supported by
  (\word{go}, {\sc past}) $\mapsto$ \word{went}) versus high-frequency related slots
  for the same lemma (e.g., (\word{think}, {\sc past\_participle}) $\mapsto$ \word{thought}
  is supported by (\word{think}, {\sc past}) $\mapsto$ \word{thought}).}

\begin{table}
\centering
	\begin{tabular}{lll} \toprule
    $\ell$ & unigram & uniform \\ \midrule
        buy & bought & \textbf{buyed} \\
        bring & brought & \textbf{bringed} \\
        feel & felt & \textbf{feeled} \\
	fight & fought &  \textbf{fighted} \\
        grind & ground & \textbf{grinded} \\
        teach & taught & \textbf{teached} \\
        hang & \textbf{hanged}$^{?}$ & \textbf{hanged}$^{?}$  \\
        think & thought & \textbf{thinked} \\
        sit & sat & \textbf{sitted}  \\
        break & broke & \textbf{breaked} \\
        see & saw & \textbf{seed} \\ \bottomrule
    \end{tabular}
    \caption{Sample output from the models trained under
    the uniform and unigram $q$ distribution at generation $t=3$. Mistakes are bolded.}
    \label{tab:model-output}
\end{table}

\subsection{Cognitive Science}\label{sec:recent-work-cogsci}
Closest to our presented work is the seminal paper of
\newcite{hare1995learning}, that first explored the paradigm of
generation learning in the context of the evolution of inflectional
morphology. They created a corpus of Old English verbs and modeled the evolution of the verbal system by successively training the neural network
\cite{rumelhart1986learning}.\footnote{We use the term neural
  network, but in terms of modern machine learning,
  \newcite{rumelhart1986learning} is best understood as a linear model
  for multi-label classification. However, contemporarily,
  the network still fit well under the moniker of connectionism. Also,
  despite attempting to solve a string-to-string task, the network
  used a static computation graph, which required an abstruse encoding
  scheme: every input was mapping to a set of Wickelphones
  \cite{wickelgren1969context}, phoneme $n$-grams.}  Our work is also closely related to \textbf{iterative language modeling},
proposed by \newcite{kirby2001spontaneous}. That work, like ours, has
a generational scheme where models teach the next generation.
Also, as in this work, they discuss the diachronic regularity and irregularity of
linguistic structure. The main difference to our work lies
in the that our model actually outputs phonological strings. Also, see \newcite{xanthos2011role} for a discussion
of children's acquisition of morphology in a variety of languages.
\subsection{Related Work in NLP}\label{sec:recent-work-nlp}
Recently, NLP has also experienced a renaissance of interest in
simulation-based approaches to language emergence. We highlight some
recent work in this area and contrast it with our proposal in this
paper.
The realm of pragmatics offers a natural setting for the exploration
of the interaction of agents in an environment. The rational speech
act (RSA) framework views pragmatics as a recursive communication
between a speaker agent and a hearer agent
\cite{frank2012predicting}. RSA has been the basis for
numerous recent simulations in NLP \cite{andreas-klein:2016:EMNLP2016}.
Other simulation work has investigated
the emergence of language in multi-agent systems.
For example, \newcite{lazaridou2016multi} focus on referential games, where agents discuss an image. They show
that neural models develop their own language, as it were,
for discussing the images. The work in this paper is similar in spirit
in that it involves neural networks communicating with each other,
while the underlying motivation is substantially different. We are interested in
the linguistic question of how language does or does not stabilize over time, rather
than whether neural models will develop language.

\section{Conclusion}
In this paper, we have revisited generational simulation of the
evolution of morphology, originally presented in the seminal work of
\newcite{hare1995learning}. Specifically, we test the hypothesis that the distribution over types determines to what degree
regularization of irregulars is possible. Different distributions lead to different stable states for the language, with varying amounts of irregularity retained.  Our simulation is
significantly updated methodologically---rather than attempting a
string-to-string transduction problem with a feed-forward network, we
use LSTM recurrent neural models. Moreover, we provide a more concrete
experimental design than present in the original study that
allows for clean hypothesis testing. We find that
(with $p < 0.002$) irregulars are more likely to die out
when we use a training distribution that underrepresents these irregular forms
or their related forms.

\section*{Acknowledgments}
The first author would like to acknowledge an \mbox{NDSEG} grant and a Facebook PhD
fellowship.  This material is also based upon work supported by the National
Science Foundation under Grant No.\@ 1718846 to the last author.

\bibliography{learnability}
\bibliographystyle{acl_natbib}

\end{document}